\pdfoutput=1
\documentclass[11pt]{article}

\usepackage[]{acl}

\usepackage{times}
\usepackage{latexsym}
\usepackage{url}

\usepackage[T1]{fontenc}
\usepackage[utf8]{inputenc}

\usepackage{microtype}

\usepackage{graphicx}
\usepackage{caption}
\usepackage{subcaption}

\title{Meeting Decision Tracker: Making Meeting Minutes with De-Contextualized Utterances}

\author{Shumpei Inoue$^1$, Hy Nguyen$^1$, Pham Viet Hoang$^1$, Tsungwei Liu$^1$, Minh-Tien Nguyen$^{2,}$\thanks{$^*$Corresponding Author.} \\
        $^1$Cinnamon AI, 10th floor, Geleximco building, 36 Hoang Cau, Dong Da, Hanoi, Vietnam. \\
        \texttt{\{sinoue, hy, hugo, tsungwei\}@cinnamon.is} \\
        $^2$Hung Yen University of Technology and Education, Hung Yen, Vietnam. \\
        \texttt{tiennm@utehy.edu.vn}}

\begin{document}
\maketitle
\begin{abstract}
Meetings are a universal process to make decisions in business and project collaboration. The capability to automatically itemize the decisions in daily meetings allows for extensive tracking of past discussions. To that end, we developed Meeting Decision Tracker, a prototype system to construct decision items comprising decision utterance detector (DUD) and decision utterance rewriter (DUR). We show that DUR makes a sizable contribution to improving the user experience by dealing with utterance collapse in natural conversation. An introduction video of our system is also available at \url{https://youtu.be/TG1pJJo0Iqo}.

\end{abstract}

\section{Introduction}

Obtaining a brief description and salient contents of meetings is a functionality that can certainly help business operations. Although automatic speech recognition enables us to transcribe meeting records automatically, its transcription is possibly much more verbose, noisy, or collapsed, and is far from being utilized in its raw form. Previous research attempted to extract important information from dialogue, such as decision-making utterances, \cite{bak2018conversational,karan2021mitigating}, extractive summaries of online forums \cite{tarnpradab2017toward,khalman2021forumsum}, or group chat threads \cite{wang2022summarization}. Another study, \citet{lugini2020discussion} presented a discussion tracker to facilitate collaborative argumentation in classroom discussion by visualizing discussion transcription.

%tarnpradab2017toward

\begin{figure*}[!h]
    \centering
    \includegraphics[width=1.0\textwidth]{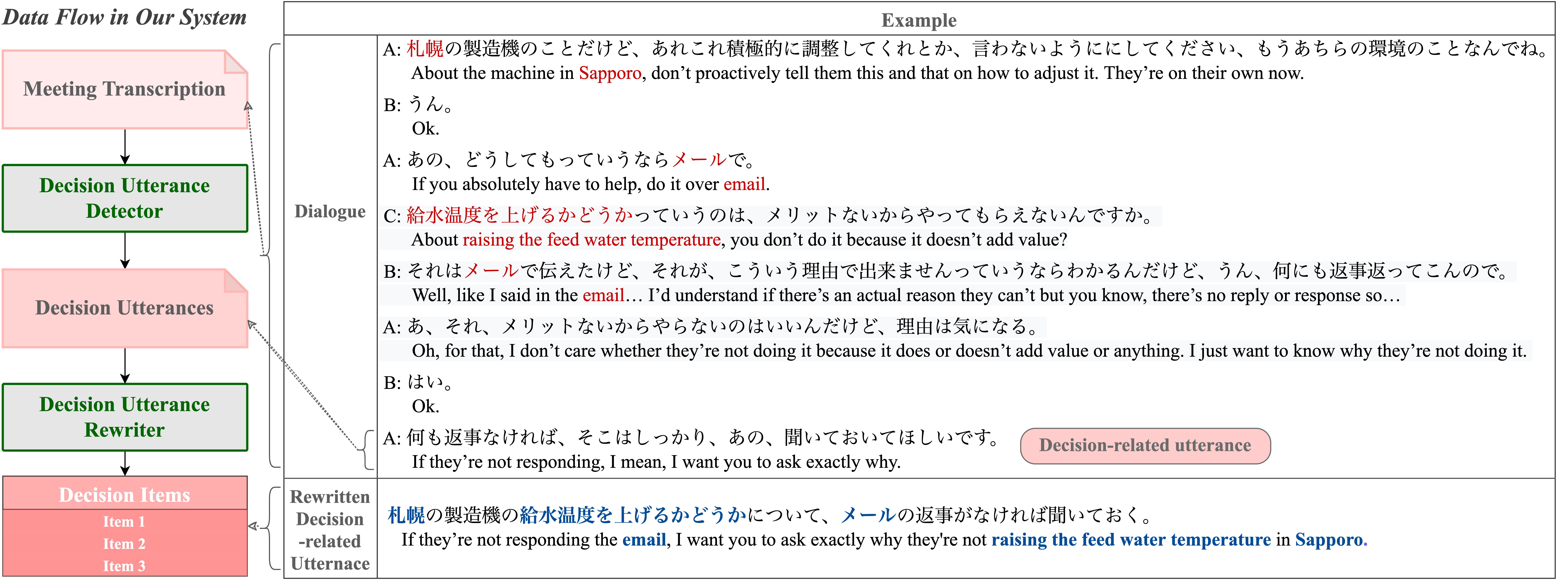}
    \caption{The data flow in our system and the conversation example. The red in the dialogue shows information omitted in the decision-related utterance. The blue shows information to be restored by Decision Utterance Rewriter.}\vspace{-0.2cm}
    \label{fig:demo_overview}
\end{figure*}

However, extracted utterances are usually incomplete and difficult to understand due to ellipses and co-references in conversations \cite{su2019improving}. Figure \ref{fig:demo_overview} (the right) shows an example of a partial dialogue ending with a decision-related utterance in our dataset. This shows that objects or indicatives in utterances in natural conversations are usually ambiguous, and the meaning of decision-related utterances has a strong dependency on context. Furthermore, especially in Japanese, the format of the spoken language is often far apart from the written language because of frank expressions and many filler phrases. This nature reduces user experience with the naive use of utterances extracted from dialogues. In response to this, Incomplete Utterance Restoration (IUR) \cite{Pan-Restoration-EMNLP-19,su2019improving,Huang-SARG-AAAI-21,inoue2022enhance} handles the problem where the model rewrites and restores incomplete utterances by considering the dialogue context with promising results. However, we have yet to see IUR models applied for practical use in actual business applications.

%Liu-RUN-BERT-EMNLP-20

This paper presents \textit{Meeting Decision Tracker (MDT)}, a system that automatically generates the itemized decision list from meeting transcription. Given the meeting transcription, MDT detects decision-making utterances and rewrites them to the \textit{de-contextualized utterance}, i.e., the written form with omissions restoration and filler removal. Such a capability allows users to look back at the previous meeting contents quickly and have asynchronous communication with no effort from a minute taker. The system has three crucial characteristics.

\begin{itemize}
    \item By combining modules for extracting and rewriting decision-related utterances, the system has a down-to-earth strategy to generate itemized decision lists from meeting transcription. The combination allows us to investigate the role of IUR in a bigger context with significant impact for real business applications.
    
    % \item The system provides two functions for decision-related utterance extraction and rewriting. The extraction extracts decision-related utterances from a dialogue and the rewriting rewrites the extracted utterances to improve user experience.
    
    % \item It is empowered by using strong methods for the extraction and writing. It allows the system to achieves high accuracy in two steps, which are applicable for actual business cases.
    
    % \item It was originally built for Japanese due to the business purpose, but, can be flexible to adapt to other languages.
    
    \item Besides the ordinary task of IUR, our rewriter handles the translation from the spoken language to written language by filtering filler phrases. It enables users to understand the decision item at a glance, which contributes to improving the user experience.
    
    \item Although our system is originally built for decision utterance itemization, the proposed method can be applied as a general solution for information extraction from the dialogue.
    
    %which combines utterance detection and rewriting,
    
\end{itemize}

\section{System Design}
The overall system architecture of Meeting Decision Tracker (MDT) is depicted in Figure \ref{fig:demo_overview} (the left). The main function of MDT is to generate decision items with de-contextualized representations from the transcription of daily business meetings. MDT comprises of two modules: Decision Utterance Detector (DUD) and Decision Utterance Rewriter (DUR). The detector extracts a decision list from meeting transcription and the rewriter translates (rewrites) the list to the written format. Figure \ref{fig:demo_overview} (the right) shows the example pair of the input and expected output for the system. The example indicates two points. First, the transcription contains decision-related utterances that can be used to summarize the content of the meeting. Second, the decision utterance itself is usually not self-consistent and comprehensible only after the utterance is restored by DUR. The next sections introduce the detector and rewriter.

%In the following subsections, we will describe the components of MDT in details.

%\subsection{Architecture}
%The user uploads to the MDT the meeting transcription composed of utterances with the indication of speaker's switch. In Figure xx, new lines in dialogue indicate the switch of speakers. The uploaded transcription is first processed by DUD to detect decision utterances. The system allows user to set the threshold for DUD where the decision utterances are detected when the likelihood of DUD is bigger than the threshold. This threshold tunes the relationship between the degree of overtaking (recall) and correctness of detected utterances (precision) according to user's requirements.

%\subsection{Use Case}

\subsection{Decision utterance detector}\label{sec:detector}
The first step of the detector is to detect decision-related utterances from transcription. We formulate the detection as a sequence labeling problem on the utterance level and describe the detector in two steps: input representation and classification.

%The user uploads to the MDT the meeting transcription composed of utterances with the indication of speaker's switch. In Figure \ref{fig:demo_overview} (the right), new lines in the dialogue indicate the switch of speakers. The uploaded transcription is first processed by DUD to detect decision utterances.

%To do that, we designed the detector based on BERT \cite{devlin-etal-2019-bert}. We also experimented with other classifiers and the results are shown in Table \ref{table:nonlin}.

\paragraph{Input representation}
The input uses the sequence of utterances $\{u_1,u_2,...,u_w\}$ for the sequential classification, where $w$ is the window size.
Following \citet{cohan2019pretrained}, we used the input representation $\{{\rm [CLS]}, u_1, {\rm [SEP]}, u_2, {\rm [SEP]},...,u_w, {\rm [SEP]}\}$, which contains the [CLS] token at the head of the whole input and [SEP] tokens at the tail for each utterance. Then the input was encoded by BERT \cite{devlin2019bert} for contextual representation. We set the window size as 5 empirically based on the observation of results.

%BERT-base model contains an encoder with 12 Transformer blocks, 12 self-attention heads, and a hidden size of 768. The input of a sequence passed to BERT should not be more than 512 tokens and output the representation of the sequence. The first token of the sequence is [CLS], containing the classification embedding, and another token [SEP] is used for separating segments. for the decision utterance detector, BERT takes the final hidden state h of the first token [CLS] as the representation of the whole input sequence

%The decision utterance detector of our approach can be formulated as a neural text classification \cite{kim-2014-convolutional}. It consists of two main modules, an unsupervised word vector and a classifier. We experiment with multiple versions of the classifier model as follows.

%\paragraph{Simple linear classifier}
%A simple and efficient baseline for text classification is to represent the text input as a bag of words (BoW) and train a linear classifier, e.g., logistic regression or an SVM \cite{10.5555/1390681.1442794}. This is the first baseline of our work.

%\paragraph{Vanilla neural classifier}
%As the second model, input tokens $\left\{x_{1}, x_{2}, \ldots x_{T}\right\}$ are passed through an unsupervised word-embedding layer \cite{bojanowski2016enriching} and a single LSTM to obtain encoded representation $h_{t}=f\left(x_{t}, h_{t-1}\right) \forall t$ for the input. The classier is a fully connected layer that uses the final encoder state $h_{t}$ to classify the correct decision at every utterance passed in.

\paragraph{Classification}
% The task for decision-related utterance extraction was formulated as a sequential sentence classification. Our model classified the tag of second utterance from the back $u_{w-2}$ considering proceeding utterances $u_1,...,u_{w-3}$ and following utterances $u_{w-1}, u_{w}$. We used the encoding of [SEP] token according to the second utterance from back as the representation $u_w$ and applied the binary classification with feedforward network. The architecture is close to \citet{cohan2019pretrained} but different in that our model does not classify the tags for proceeding utterances. In the training phase, the pre-trained weights of BERT were frozen and the parameters of the feedforward network were set to learnable.

There are several studies have addressed the decision utterance detection as a classification. \citet{Fernandez2008-ac} defined the decision-making sub-dialogues as being composed of several dialogue act tags such as the introduction of issue, decision adopted/proposed/confirmed, agreement. \citet{murray2008detecting} created abstract describing decisions, actions and problems of meeting and then associated the utterances used for abstract as the action item utterances.
\citet{chen2016aimu} classified action items in the token level following the semantic intent schema.

In this study, the task of decision-related utterance extraction was formulated as binary sequence labeling on the sentence level, different from \citet{Fernandez2008-ac}. This is because we want to keep a simple setting to confirm the efficiency of IUR in actual cases. To take advantage of context, we followed \citet{cohan2019pretrained} to jointly encode consecutive utterances. Preceding utterances leading to decision are essential because followed by \citet{Fernandez2008-ac}, we hypothesize that the particular kinds of patterns of conversation co-occur with decision. Utterances following decision are also important since affirmative response by others supports the confidence of detection.

%who focused on six classes: issue, decision adopted / proposed / confirmed, agreement.

For sequence labeling, the model uses the encoding of [SEP] tokens corresponding to each utterance and predicts tags (decision or not) by a feedforward network. Different from \citet{cohan2019pretrained}, we used only the prediction for the second utterance from the back in the input and slide the window with the stride of 1 over conversation to obtain the predictions for all utterances.

\subsection{Decision utterance rewriter}
After extracting decision-related utterances, the rewriter translates the extracted utterances from the spoken to written language to improve user experience. We describe the rewriter in two steps: input representation and rewriting.
% We adopted the incomplete utterance restoration (IUR) model \cite{inoue2022enhance} due to the same purpose of rewriting incomplete utterances.
%We adopted JET \cite{inoue2022enhance} as rewriter based on the promising performance in our experiments (Section \ref{par:dur}).

\paragraph{Input representation}
The input of DUR comprises of utterances $\{u_1,u_2,..., u_n\}$ where $u_1, ..., u_{n-1}$ is contextual dialogue and the tail utterance $u_n$ is the decision-related utterance. For input representation, we followed \citet{inoue2022enhance} to use three types of special tokens, $[X1]$, $[X2]$ and $\verb|<|\verb|\|s\verb|>|$. We inserted $[X1]$ after each utterance in contextual dialogue $u_i$ for $i=1,..,n-1$, $[X2]$ after decision-related utterance $u_n$, and $\verb|<|\verb|\|s\verb|>|$ at the tail of whole input as the EOS token.
For inference, DUR rewrites only the decision utterances detected by DUD. For each decision utterance, we used preceding utterances, including up to 360 tokens by the T5's tokenizer as the contextual dialogue.

\paragraph{Rewriting}
JET \cite{inoue2022enhance} was adopted and fine-tuned on our dataset for utterance rewriting. JET uses T5 \cite{raffel2020exploring} for the picker and writer which were jointly trained for picking important tokens and text generation. The picker picks up important tokens from dialogue context which contribute to rewriting. The two components are jointly optimized by sharing parameters of the T5's encoder, which allows the model to restore omitted information while keeping the capability of abstractive text generation to translate from the spoken to written form with fillers removal.

%The objective of writer is the rewriting for de-contextualization while the objective of picker is to pick up important tokens in dialogue context, the set of tokens omitted due to coreference and ellipse in todo-related utternace, as a sequence tagging task. The set of important tokens are automatically built by their heuristics. Since JET keeps the T5 architecture as rewriter, it can handle the translation from spoken to written form and deleting fillers as the text generation task as well as IUR's restoration.

\section{Evaluation}
In this section, we first show data annotation for the detector and rewriter, and then describe the settings used for experiments. We finally report the results and discussion of the detector and rewriter.

\subsection{Dataset}\label{sub:dataset}
%Explanation how to construct the dataset, basic statistic of data.

\paragraph{Decision utterance detector}\label{par:dud}
Our Japanese dataset was constructed based on multi-party conversations with various users' intents and decisions in real-world business scenarios. We recorded client meetings in a variety of fields, including banking, finance, and insurance, and accurately transcribed all speeches including fillers. 

For decision detection annotation, as stated in Section \ref{sec:detector}, we adopted the schema of binary to decide whether an utterance is a decision (labeled by TD) or not (non-TD). With this simple schema, we aimed to extract decision-related utterances with high coverage and relied on rewriter to restore the contextual information involving decision.
% , unlike decision sub-dialogue extraction \cite{Fernandez2008-ac}.

To do the annotation, we asked three annotators who have at least N2 Japanese skills to give a label for each utterance whether it is a decision utterance or not. N2 Japanese members are those who have ability to understand Japanese used in everyday situations and in a variety of circumstances to a certain degree.\footnote{https://www.jlpt.jp/e/about/levelsummary.html} We combined three annotators to create three groups in which each group has two annotators. To reduce resources and avoid specific bias, each group was assigned a small part of the dataset for annotation. To maintain label quality, annotators prepared a list of the specific expressions frequently used in decision utterances such as "I decided to...", "I have to..." and shared it between them. It comes from the observation that utterances containing the specific expression tend to be decision-related utterances. Each utterance was tagged by two annotators and if the tags differed, the final tag was determined after reconsideration. The Cohen Kappa agreement computed over the three groups is 0.672, showing that the agreement is moderate. It is understandable because transcription is quite noisy compared to common data types, e.g., news. The annotated data was divided into training, validation, and testing sets by meeting units and contains 27006, 3030, and 1425 utterances. The dataset is highly imbalanced where decisions only account for 6\% of the entire data, creating challenges for classifiers.

\paragraph{Decision utterance rewriter}
We created the dataset for DUR based on the DUD dataset. We selected 1120 utterances tagged by \textbf{TD} and extracted their preceding utterances containing up to 360 tokens. Two native Japanese annotators created the rewritten version of decision utterances. Annotators re-wrote decision utterances with three requirements: (i) restore omitted information extracted from preceding utterances, (ii) remove fillers, and (iii) convert from the spoken form to written form. To prepare a consistent dataset, annotators reused the original words in contextual dialogue for rewriting as much as possible, rather than creating new phrases. Annotators also checked rewriting each other every 100 samples to align the quality.

\subsection{Experimental settings}
For the detector, we used pretrained BERT \cite{cohan2019pretrained} (cl-tohoku/bert-base-japanese) and fine-tuned MLP (dimensions $512$, $400$, $5$) by AdamW in $20$ epochs with drop-out of $0.2$, the batch size of $16$, and the learning rate of $5e-5$. For rewriter, we trained JET with pretrained t5-base-japanese T5 by AdamW with weight decay of $0.01$ in $70$ epochs with the batch size of $6$, the leaning rate of $2e-5$, and the beam size of $5$. All models were trained on a single Tesla P100 GPU.

\subsection{Results and discussion}
    
\paragraph{Decision utterance detector} \label{par:dud}
%\textbf{Logistic regression (LR)} uses bag of words (BoW) features for classification \cite{10.5555/1390681.1442794}.
%\cite{srivastava2014dropout}
%\cite{bojanowski2016enriching}

We compared the BERT model with two different task formulations: sequential sentence labeling (SL) and sentence classification (SC). For sequence labeling, we used the same model described in Section \ref{sec:detector}. For sentence classification, we trained the model by using BERT to predict the tag of the second utterance from the back given the input utterances \{$u_1$,$u_2$,...,$u_w$\}. It follows input representation in Section \ref{sec:detector} and uses the [CLS] tokens for binary classification.
To deal with the imbalanced dataset, we also tested the model with {\bf back translation} (BT), a technique to augment the data by translating original text data into another language and then back into the original language. We augmented the positive samples\footnote{positive sample refers the consecutive utterances $u_1, ..., u_{w}$ with the decision tags for $u_{w-2}$.} by seven times using seven languages.\footnote{We used Google Translate API with 7 languages, "vi", "en", "zh-CN", "zh-TW", "fr", "de", "ko"}

% We applied the back translation technique to deal with the imbalanced data. {\bf Back translation }(BT) \cite{edunov2018understanding} is a technique to augment the data by translating text data into another language and then back into the original language. We augmented the positive samples by seven times using seven languages.

% We compared the detector based on BERT with its version for dealing with imbalanced data. We also implemented a single layer \textbf{LSTM} uses unsupervised word vectors with dropout to obtain the encoded representation of inputs for classification. \textbf{BERT} is the model described in section \ref{par:dud}. 
% We applied the back translation technique to deal with the imbalanced data. {\bf Back translation }(BT) \cite{edunov2018understanding} is a technique to augment the data by translating text data into another language and then back into the original language. We augmented the positive samples by seven times using seven languages.
%{\bf focal loss} \cite{lin2017focal} is simple but effective method for data imbalance problem that down-weights the loss assigned to examples well-classified. 

%{\bf upsampling} (up) randomly used positive samples\footnote{positive sample refers the consecutive utterances $u_1, ..., u_{w}$ with the decision tags for $u_{w-2}$.} (data with decision tag) delicately to virtually tune the ratio of positive and negative same.

\begin{table}[!h]
\caption{Results of the Decision utterance detector.}
\centering 
\begin{tabular}{l c c c} % centered columns (4 columns)
\hline %inserts double horizontal lines
Method & Precision & Recall & F1 \\ % inserts table
%heading
\hline % inserts single horizontal line
BERT (SC) & 0.32 & \textbf{0.59} & 0.42 \\
BERT (SC) + BT & 0.33 & 0.58 & 0.42 \\
BERT (SL) & \textbf{0.48} & 0.55 & \textbf{0.51} \\
BERT (SL) + BT & 0.44  & 0.55  & 0.49  \\
\hline %inserts single line
\end{tabular}
\label{table:dud} % is used to refer this table in the text
\end{table}
Table \ref{table:dud} shows the performance comparison. 
As we can observe, sequence labeling (BERT (SL)) without using back translation is the best. BERT (SL) achieves better performance than BERT (SC) in general. This suggests that the knowledge of jointly predicting tags helps to better understand the dependencies between utterances. So it leads to improving the performance. Binary sentence classification does not show high F-scores even though the model uses context by using concatenation. It suggests more sophisticated combinations for improving the performance of binary sentence classification. Back translation does not help to improve the quality of the detector. This is because utterances are quite broken in terms of writing and contain fillers. It suggests other data augmentation methods for conversation.

\paragraph{Decision utterance rewriter} \label{par:dur}
For the writing part, we compared JET to T5 \cite{raffel2020exploring} and s2s-ft \cite{bao2021s2s} due to its efficiency for the IUR task. {\bf T5} uses a text-to-text framework pre-trained on data-rich tasks with transformer encoder-decoder. {\bf s2s-ft} applies attention masks with fine-tuning methods for the generation task. We did not report the results of ProphetNet \cite{qi2020prophetnet} and UniLM \cite{dong2019unified} due to no pre-trained models for Japanese; SARG \cite{Huang-SARG-AAAI-21} and RUN-BERT \cite{Liu-RUN-BERT-EMNLP-20} due to its low accuracy for IUR \cite{inoue2022enhance}.

%This is because for ProphetNet and UniLM, there is no pre-trained weight for Japanese text generation, and for SARG and RUN-BERT, they are too specialized for the IUR task to be applicable for abstractive generation as described in \citet{inoue2022enhance}. We applied beam search with the beam size of 5 for all models when decoding.

%conducted the experiment with three representative text generation models: JET \cite{inoue2022enhance}, T5 \cite{raffel2020exploring}, and s2s-ft \cite{Bao2021-ki}. {\bf JET} optimizes rewriting task utilizing T5 with auxiliary loss that deals with information omission by sequence tagging. 

For evaluation, we followed \citet{Pan-Restoration-EMNLP-19} to use ROUGE, BLEU and f-scores.\footnote{We used sumeval for ROUGE and BLEU scores (https://github.com/chakki-works/sumeval) and f-scores are based on $n$-grams with the MeCab tokenizer.} All methods used the beam width of 5. To obtain the reliable comparison, we also report the human evaluation by using {\bf Text Flow} and {\bf Understandability} \cite{Kiyoumarsi2015-ul}. {\bf Text Flow} shows how the rewritten utterance is correct grammatically and easy to understand. {\bf Understandability} shows how much the prediction is similar to reference semantically. Three annotators (who are at least N2 Japanese skills) involved the judgement and each annotator gave a score (1: bad; 2: acceptable; 3: good) to each rewritten utterance. The three evaluators scored for each 190 testing samples and the final scores were calculated by the average of scores from the evaluators.

%proposed by \citet{Kiyoumarsi2015-ul} with 3 grades (1: bad; 2: acceptable; 3: good): {\bf Text Flow} shows how the rewritten utterance is correct grammatically and easy to understand. {\bf Understandability} shows how much the prediction is similar to reference semantically. Our three evaluators scored for each 190 test samples and the final scores were calculated taking average. Both Table \ref{table:dur} and \ref{table:humaneval} show JET is consistently the best over all metrics. Consequently, we adopted JET as our rewriter module.

\begin{table}[!h]
\centering
\setlength{\tabcolsep}{3.5pt}
\caption{Results of Decision utterance rewriter. RG is ROUGE and BL stands for BLEU.}\label{tab:dur-results}
\begin{tabular}{clccccc} \hline
& Method & RG-1 & RG-2 & BL & f1   & f2     \\ \hline
    & JET   &{\bf 56.71}    &{\bf 36.60}    &{\bf 25.97}   &{\bf 36.81}    &{\bf 21.52}  \\
    & T5   &54.91   &35.10    &24.48    &36.61    &21.42     \\
    & s2s-ft & 47.71   &29.52    &19.91    &27.41    &15.74  \\ \hline
\end{tabular} \label{table:dur}
%\end{table}

% \begin{table}[!h]
\centering
\setlength{\tabcolsep}{5pt}\vspace{0.2cm}
\caption{Human Evaluation}
\begin{tabular}{clcc} \hline
& Method & Text Flow &Understandability     \\ \hline
    & JET  &{\bf 2.53} &{\bf 1.90} \\
    & T5    &2.41 &1.79 \\
    & s2s-ft  &2.16 &1.55  \\ \hline
\end{tabular}\label{table:humaneval}
\end{table}

Results in Tables \ref{table:dur} and \ref{table:humaneval} show that JET is the best for both automatic and human evaluation. This is because the model was empowered by T5 and the picker, that picks up important tokens for rewriting. T5 is the second best due to the strong pre-trained model for Japanese. s2s-ft does not show competitive performance compared to model with text-to-text pre-training framework. 

% s2s-ft does not show good performance due to the ability of transformers (e.g. BERT) for the generation task.
\paragraph{Effectiveness of utterance rewriter}
A human evaluation was conducted to see how the rewriter contributes to improving the quality of decision items. A good rewriter requires (i) to keep the original contents before writing and (ii) to enrich the content by supplementing omitted information. Given the pair of the original decision utterance (ODU) and the rewritten decision utterance (RDU), we defined the scoring criteria in the range of 1 to 5 as the following.
\begin{enumerate}
  \setlength{\parskip}{0cm} % 段落間
  \setlength{\itemsep}{0cm} % 項目間
    \item RDU completely lost meaning of ODU.
    \item RDU somewhat lost meaning of ODU.
    \item RDU keep meaning of ODU but no additional information.
    \item RDU keep meaning of ODU with a few additional information.
    \item RDU keep meaning of ODU with sufficient additional information.
\end{enumerate}
% Rewriting lost original content of decision utterance completely (score 1) or partially (score 2). Rewriting keep the original content and have no (score 3) / a few (score 4) / sufficient (score 5) additional information. 
As far as the RDU lost the meaning of ODU, the score would be 1 or 2 even there was any additional information. In accordance with this criteria, we collected the scores from the three evaluators by using the utterances before and after the rewriting on test data from the DUR dataset. These three evaluators are annotators who also worked to construct the RDU dataset (Section \ref{sub:dataset}).
\begin{table}[!h]
\centering
\setlength{\tabcolsep}{5pt}
\caption{Effectiveness evaluation.}
\begin{tabular}{lcccccc} \hline
    score & 1 & 2 & 3 & 4 & 5     \\ \hline
    ratio  &3.76\% & 7.04\% & 20.7\% & 27.7\% &40.8\%  \\ \hline
\end{tabular}\label{table:rewriter-eval}
\end{table}

Table \ref{table:rewriter-eval} shows the result of evaluation with the ratio of each score. It indicates 10.8\% of samples decrease in quality (score $\leq 2$) while  68.5\% of samples increase in quality (score $\geq 4$). The average score from the evaluators was 3.948, higher than 3, showing that the quality of decision items increases by our rewriter in general. These results show our rewriter certainly contributes to better user experience when displaying decision items (Figure \ref{fig:demo2}).

\section{Demonstration Scenario}
We provide a UI\footnote{The system: https://bit.ly/3sH6193; user and pwd are \texttt{Guest123@MDT.com}. Please skip verification when login.} that allows users to look back at decision-making items in past meetings at a glance. Especially in business settings, the accumulation of daily meetings can be compactly stored as itemized decisions to be accessed easily and support project progress management and sharing.

\begin{figure}[!h]
  \centering
  \begin{subfigure}{0.56\textwidth}
    \centering
    \includegraphics[width=80mm]{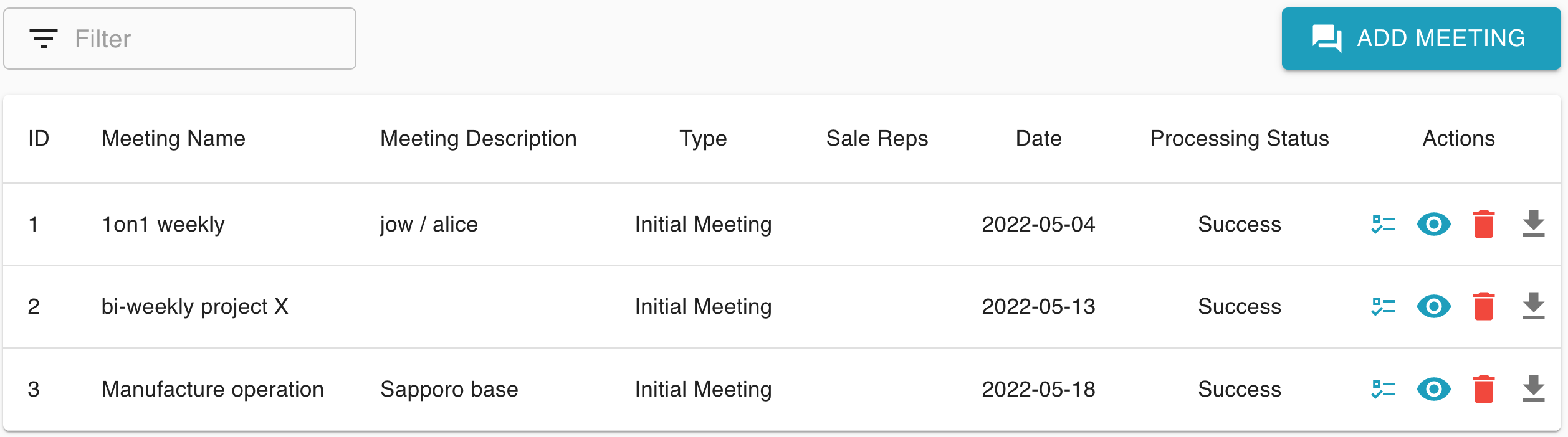}
    \caption{The uploaded meeting list.}
    \label{fig:demo1}
  \end{subfigure}
  
   \begin{subfigure}{0.56\textwidth}
    \centering
    \includegraphics[width=80mm]{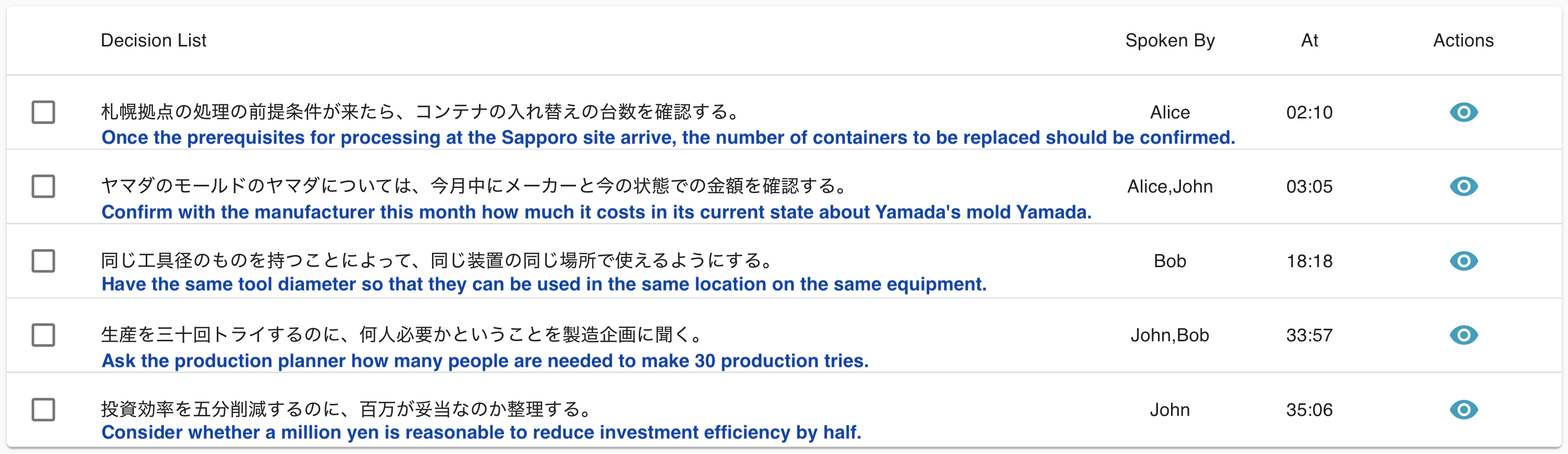}
    \caption{The decision items.}
    \label{fig:demo2}
  \end{subfigure}
  
  \begin{subfigure}{0.56\textwidth}
    \centering
    \includegraphics[width=80mm]{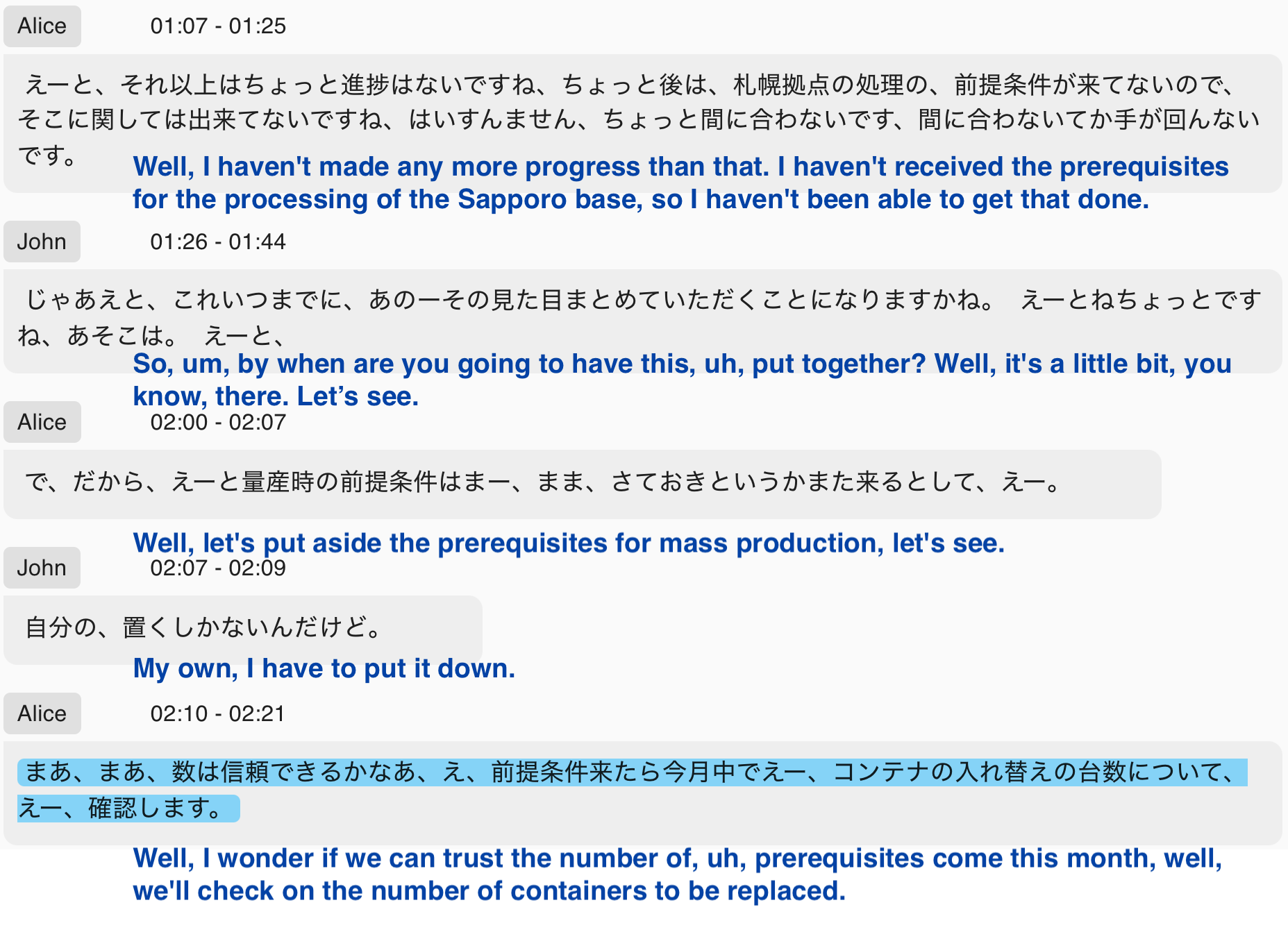}
    \caption{The original transcription.}
    \label{fig:demo3}
  \end{subfigure}
  \caption{The screenshots from the system.}\vspace{-0.3cm}
  \label{fig:demo}
\end{figure}

Figure \ref{fig:demo} shows an example processed by our system. The original decision-related utterance is highlighted in blue in Figure \ref{fig:demo3}. Its content \textit{"Well, I wonder if we can trust the number of, uh, prerequisites come this month, well, we'll check on the number of containers to be replaced."} is rewritten and displayed in the first line of the decision list in Figure \ref{fig:demo2} as the de-contextualized form, \textit{"Once the prerequisites for processing at the Sapporo site arrive, the number of containers to be replaced should be confirmed.."}.

The view of the first run is the list of the meetings uploaded (Figure \ref{fig:demo1}). When users click on the meeting from the list they intend to go back, so decision items for corresponding meeting are unfolded (Figure \ref{fig:demo2}). Here we display \textit{de-contextualized} decision by DUR instead of original decision-related utterances. Since the DUR module makes decision-related utterances self-contained in the written language format, displayed decision items are straightforward and user-friendly to quickly understand them. To allow users to see the context of the discussion, users can click on the decision item and view the original transcription with a scrolling position where the corresponding decision-related utterance is at the bottom (Figure \ref{fig:demo3}).

% \begin{figure*}[htbp]
% \centering
% \subfloat[subtitle-A]{\includegraphics[clip, width=80mm]{figs/screen1.png}
% \label{fig:label-A}}
% \\
% \subfloat[subtitle-B]{\includegraphics[clip, width=80mm]{figs/screen2.png}
% \label{fig:label-B}}
% \\
% \subfloat[subtitle-C]{\includegraphics[clip, width=80mm]{figs/screen3.png}
% \label{fig:label-C}}

% \caption{Figures-ABC.}
% \label{fig:label-ABC}
% \end{figure*}

% \begin{figure}[htbp]
%     \begin{minipage}{0.5\hsize}
%         \begin{center}
%         \includegraphics[width=80mm]{figs/screen1.png}
%         \caption{first fig}
%         \label{fig:one}
%         \end{center}
%     \end{minipage}%

%     \begin{minipage}{0.5\hsize}
%         \begin{center}
%         \includegraphics[width=80mm]{figs/screen2.png}
%         \caption{second fig}
%         \label{fig:two}
%         \end{center}
%     \end{minipage}
    
%     \begin{minipage}{0.5\hsize}
%         \begin{center}
%         \includegraphics[width=80mm]{figs/screen3.png}
%         \caption{second fig}
%         \label{fig:two}
%         \end{center}
%     \end{minipage}
% \end{figure}

\section{Conclusion and Future Work}
In this paper, we presented \textit{Meeting Decision Tracker}, a system to automatically itemise the decision-making in daily meetings as well as the tracking of past discussions. We showed the effective adaptation of IUR for decision-item tracking in the context of actual business scenarios. MDT not only displays itemized decision-utterances with an easy-to-understand format, but also allows users to go back and review the contextual dialogue deriving for the decision.
Future work will firstly improve the quality of the detector and rewriter. Other potential directions will incorporate ASR into MDT to create an end-to-end system and add functions to remind users of detected decisions and to search for past meetings.

\section*{Acknowledgement}
We would like to thank Nguyen Duy Anh for the discussion and support of the decision utterance detector. We also thank anonymous reviewers who gave constructive comments on our paper.

%by dealing with omissions and coreference, which are inevitable in conversation

\bibliography{ref}
\bibliographystyle{acl_natbib}

\end{document}